\documentclass[12pt]{article}
\usepackage{amsmath}
\usepackage{amsfonts}
\usepackage{times}
\usepackage{graphicx}
\usepackage{graphics}
\usepackage{color}
\usepackage{multirow}
\usepackage{caption}
\usepackage{subcaption}
\usepackage[
backend=biber,
style=apa,
maxcitenames=1
]{biblatex}
\usepackage{rotating}
\usepackage{bbm}
\usepackage{latexsym}
\DeclareMathOperator*{\argmax}{arg\,max}
\usepackage{soul}

\newcommand{\captionfonts}{\normalsize}

\makeatletter  
\long\def\@makecaption#1#2{%
  \vskip\abovecaptionskip
  \sbox\@tempboxa{{\captionfonts #1: #2}}%
  \ifdim \wd\@tempboxa >\hsize
    {\captionfonts #1: #2\par}
  \else
    \hbox to\hsize{\hfil\box\@tempboxa\hfil}%
  \fi
  \vskip\belowcaptionskip}
\makeatother   

\usepackage{lineno}

\begin{document}
\hspace{13.9cm}1

\ \vspace{20mm}\\

\begin{center} {\LARGE Training a HyperDimensional Computing Classifier using a Threshold on its Confidence} \end{center}
\ \\
{\bf \large Laura Smets$^{\displaystyle 1}$, Werner Van Leekwijck$^{\displaystyle 1}$, Ing Jyh Tsang$^{\displaystyle 1}$, Steven Latr\'e$^{\displaystyle 1}$}\\
{$^{\displaystyle 1}$University of Antwerp - imec, IDLab - Department of Computer Science, Sint-Pietersvliet 7, 2000 Antwerp, Belgium, E-mail: \{Laura.Smets,Werner.Vanleekwijck\}@uantwerpen.be; \{Inton.Tsang,Steven.Latre\}@imec.be.}\\
%

{\bf Keywords:} Hyperdimensional Computing, Vector Symbolic Architectures

%
%
\begin{center} {\bf Abstract} \end{center}
Hyperdimensional computing (HDC) has become popular for light-weight and energy-efficient machine learning, suitable for wearable Internet-of-Things (IoT) devices and near-sensor or on-device processing. HDC is computationally less complex than traditional deep learning algorithms and achieves moderate to good classification performance. This article proposes to extend the training procedure in HDC by taking into account not only wrongly classified samples, but also samples that are correctly classified by the HDC model but with low confidence. As such, a confidence threshold is introduced that can be tuned for each dataset to achieve the best classification accuracy. The proposed training procedure is tested on UCIHAR, CTG, ISOLET and HAND dataset for which the performance consistently improves compared to the baseline across a range of confidence threshold values. The extended training procedure also results in a shift towards higher confidence values of the correctly classified samples making the classifier not only more accurate but also more confident about its predictions.

\section{Introduction}
Hyperdimensional computing (HDC) has gained a lot of interest in the field of low-power, brain-inspired artificial intelligence (AI). It tries to mimic the human brain by distributing the information across thousands of vector elements in analogy to the large number of neurons present in our brains. HDC is a light-weight and energy-efficient algorithm that has already been used in several applications which can be divided in three categories according to \textcite{Ge2020}: (i) text classification (\cite{Rachkovskij2007}; \cite{Rahimi2016b}), (ii) signals such as speech recognition (\cite{Imani2017}), human activity recognition (\cite{Kim2018}), handgesture recognition (\cite{Rahimi2016a, Moin2021, Zhou2021}) and time series classification (\cite{Schlegel2022}), and (iii) images such as classification of medical images (\cite{Kleyko2017a, Watkinson2021}), character recognition (\cite{Manabat2019}) and robotics (\cite{Neubert2019}). It has been shown that HDC is suitable for wearable Internet-of-Things (IoT) devices, near-sensor AI applications and on-device processing due to few data requirement (\cite{Rahimi2019}), robustness to noise (\cite{Kanerva2009,Widdows2015,Rahimi2019}), low latency (\cite{Rahimi2019}) and fast processing (\cite{Rahimi2019}). This avoids the limitations of IoT architectures in which data is sent to the cloud and consequently processed causing high latencies, large communication energy and privacy concerns (\cite{Basaklar2021}). \\

\noindent Although HDC has the advantage of being computationally less complex than traditional deep learning algorithms, it is only able to achieve moderate to good performance in classification tasks. Hence, research is ongoing to adjust and improve HDC to boost its performance. This article aims to contribute to this research by proposing a simple, yet effective extended training procedure in the binary HDC framework to improve its performance on signal applications, suitable for wearable IoT devices. The next section gives a detailed introduction to HDC after which HDC adjustments that are already proposed in literature are discussed in Section 3. Thereafter, the proposed extended training procedure is introduced in Section 4. In the fifth section, an overview of the performed experiments is given of which the results are presented and discussed in the sixth section. Finally, the conclusions of the article are presented.

\section{Hyperdimensional Computing}

HDC is a mathematical framework using hyperdimensional (HD) vectors (i.e., vectors with very high dimension typically up to ten thousands, also called hypervectors (HVs)) and simple HD arithmetic operations to represent data. The focus of this article is on dense binary HVs (i.e., the elements are 0 or 1 with an equal probability of occurrence of both values). The analysis of data relies on the similarity between HVs which is calculated using the normalized Hamming distance between two binary HVs $\textbf{v}_{1}$ and $\textbf{v}_{2}$\footnote{\normalsize A list of used symbols can be found in the appendix (Table \ref{tab:symbols}).}:
\begin{equation} \label{eq:similarity}
    s(\textbf{v}_{1},\textbf{v}_{2}) = 1 - \frac{h(\textbf{v}_{1},\textbf{v}_{2})}{D}
\end{equation}
with $s$ the similarity between $\textbf{v}_{1}$ and $\textbf{v}_{2}$, $D$ the dimensionality (e.g., $D = 10,000$) and $h$ the Hamming distance between $\textbf{v}_{1}$ and $\textbf{v}_{2}$ which counts the number of elements for which the coordinates differ (i.e., the sum of elements of the exclusive disjunction (XOR)):
\begin{equation} \label{eq:hamming}
    h(\textbf{v}_{1},\textbf{v}_{2}) = \sum_{d = 1}^{D} (\textbf{v}_{1}[d] \textrm{ XOR } \textbf{v}_{2}[d]).
\end{equation}
The HD arithmetic operations include: \\
\textbf{(a) bundling} $\oplus$: $\mathcal{B} \times \mathcal{H} \to \mathcal{B}$: $(\textbf{B},\textbf{v}) \to \textbf{B}+\textbf{v}$ where $\mathcal{B} = \mathbb{N}^{D}$ and $\mathcal{H} = \{0,1\}^{D}$ (i.e., element-wise addition) after which the bundle $\textbf{B}$ is binarized into the HV $\textbf{v}$ with the majority rule $[.]: \mathcal{B} \to \mathcal{H}$: $\textbf{B} \to \textbf{v}$ according to:
\begin{equation} \label{eq:majority}
    \textbf{v}[d] = [\textbf{B}[d]] =
    \begin{cases}
        1         & \text{if} \, \textbf{B}[d] > \frac{n}{2} \\
        0         & \text{if} \, \textbf{B}[d] < \frac{n}{2} \\
        rand(0,1) & \text{if} \, \textbf{B}[d] = \frac{n}{2}
    \end{cases}
\end{equation}
with $n$ the number of HVs bundled in $\textbf{B}$ and $rand(0,1)$ means that the component $\textbf{v}[d]$ is randomly assigned to 0 or 1 in the presence of ties (which can only occur when the number of bundled vectors is even); \\
\textbf{(b) binding} $\otimes$: $\mathcal{H} \times \mathcal{H} \to \mathcal{H}$: $(\textbf{v}_{1},\textbf{v}_{2}) \to \textbf{v}_{1} \textrm{ XOR } \textbf{v}_{2}$ (i.e., XOR in binary HDC); and \\
\textbf{(c) permutation} $\rho$ (i.e., cyclic shift in binary HDC). \\

\begin{figure}[h]
\centering
\includegraphics[width=.7\linewidth]{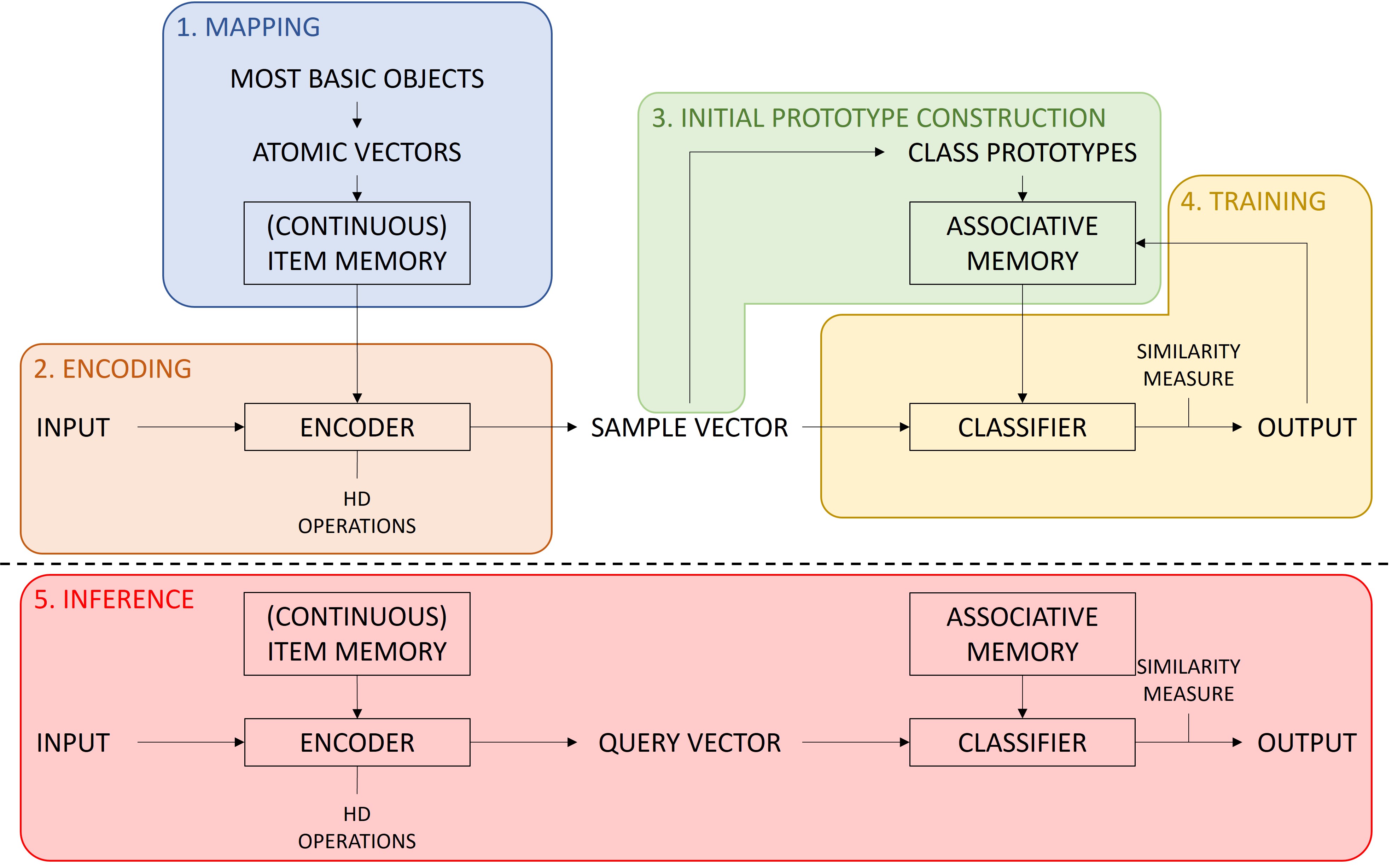}
\caption{Schematic overview of the HDC framework in which five main steps are distinguished: (1) mapping, (2) encoding, (3) initial prototype construction, (4) training and (5) inference.}
\label{fig:overview}
\end{figure}

\noindent Figure \ref{fig:overview} gives a schematic overview of the framework of HDC in which five main steps are distinguished: (1) mapping, (2) encoding, (3) initial prototype construction, (4) training and (5) inference, which will be explained in more detail in the following section. \\

\begin{figure}[h]
\centering
\includegraphics[width=.5\textwidth]{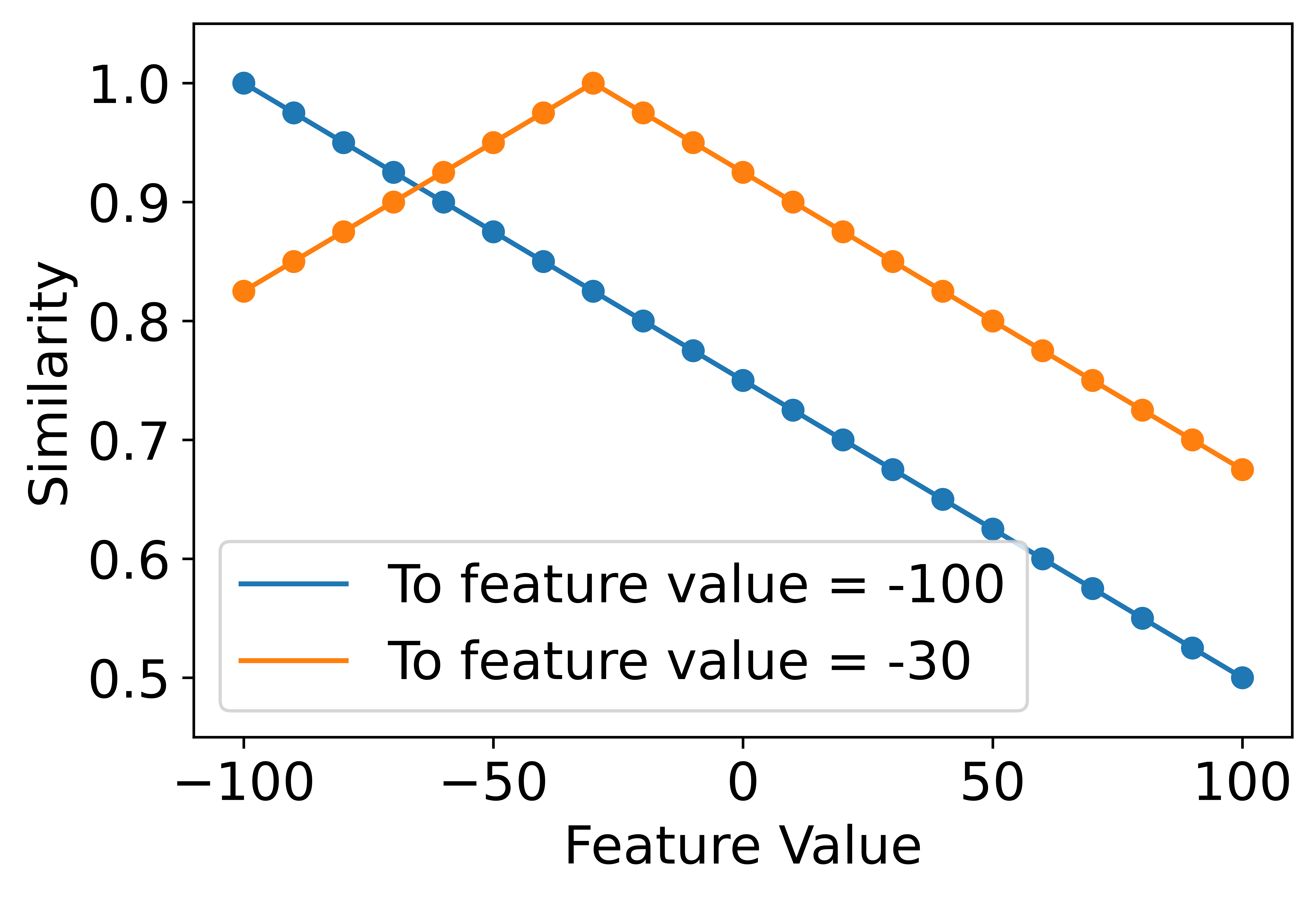}
\caption{Example of \textit{linear mapping} for a feature with discrete values ranging from -100 to 100 with steps of 10. The similarity of each feature value's level hypervector to the lowest level hypervector (feature value $= -100$) and to the hypervector for the feature value of $-30$ is shown.}
\label{fig:linearmapping}
\end{figure}

\noindent \textbf{(1) Mapping.} The way of mapping  depends on the type of data: \\
\textbf{(a)} For nominal data, each possible category is mapped to a randomly chosen atomic HV and stored in an Item Memory (IM). These random HVs are pseudo-orthogonal in high dimensional spaces which converges to exact orthogonality with increasing dimensionality (\cite{Kleyko2022a}). \\
\noindent \textbf{(b)} In the case of ordinal or discrete data, there is a natural ordering of levels of categories or integer values such that closer levels should be mapped to more similar HVs than levels further apart. This is typically achieved by a Continuous Item Memory (CIM) applying \textit{linear mapping} of levels to atomic HVs (\cite{Rahimi2016a,Kleyko2018b}). Figure \ref{fig:linearmapping} illustrates the similarity of values to the lowest level (feature value = $-100$) that decreases linearly up until orthogonality (similarity = 0.5) and the similarity of values to the feature value equal to $-30$ that decreases linearly for smaller and larger feature values. \\

\noindent \textbf{(c)} Continuous data is quantized with a quantization step into a predefined number of discrete levels to which \textit{linear mapping} can be applied. \\

\begin{figure}[h]
\centering
\includegraphics[width=.8\linewidth]{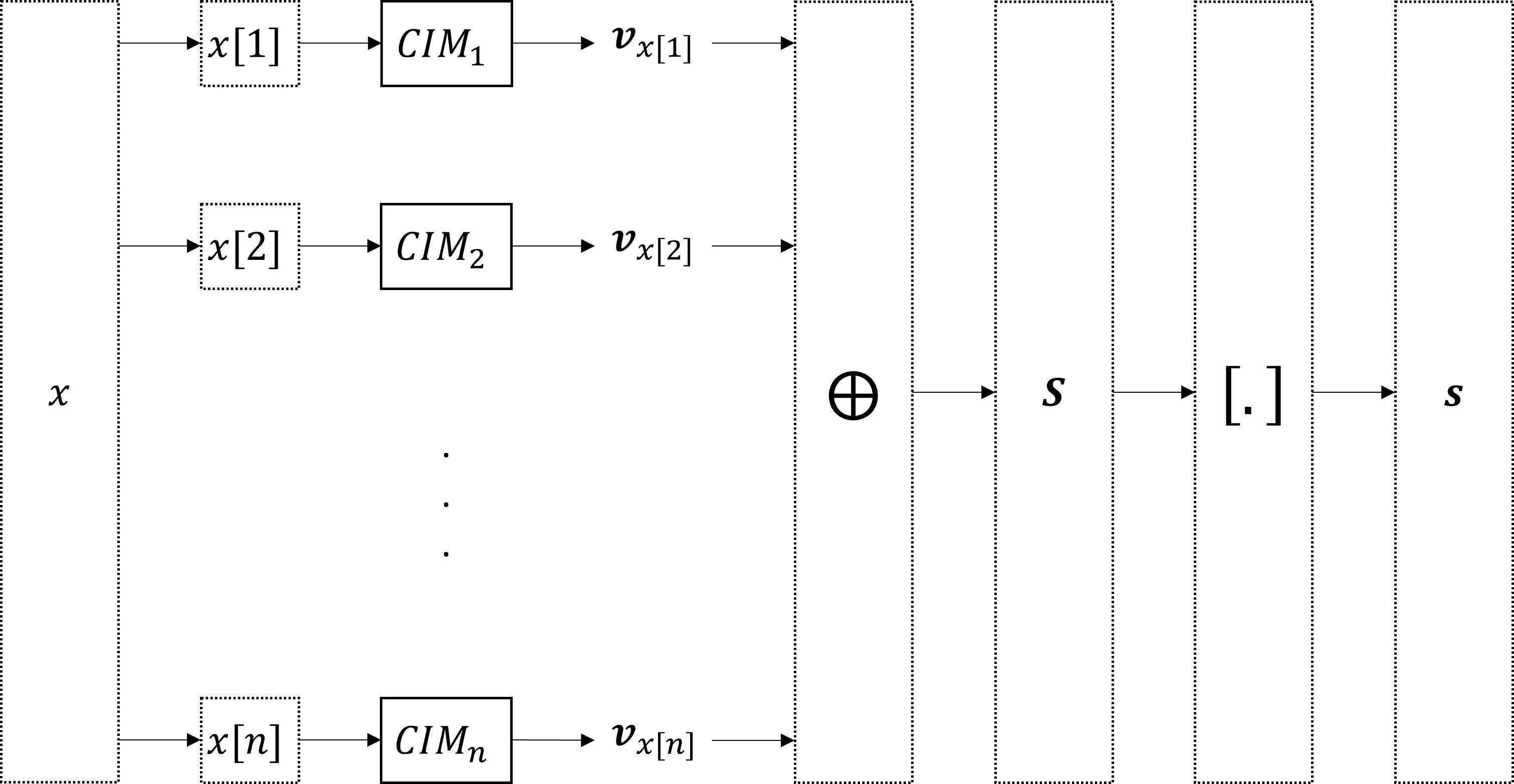}
\caption{Schematic overview of the encoding with a separate CIM for the values of each feature.}
\label{fig:spatial}
\end{figure}

\noindent \textbf{(2) Encoding.} Input data is encoded in HVs using the atomic vectors made in the previous step, and HD arithmetic operations. An input sample $x$ having $n$ features is encoded as (Figure \ref{fig:spatial}): \\
For each feature ($j = 1...n$), a CIM translates the feature value to an HV $\textbf{v}_{x[j]}$. Next, all $\textbf{v}_{x[j]}$'s are bundled together to form the sample bundle $\textbf{S}$ by initializing
\begin{equation} \label{eq:samplesa}
    \textbf{B}_{0} = \{0\}^D
\end{equation}
and bundling each $\textbf{v}_{x[j]}$ one at a time:
\begin{equation} \label{eq:samplesb}
    \textbf{B}_{j} = \textbf{B}_{j-1} \oplus \textbf{v}_{x[j]}.
\end{equation}
The sample bundle \textbf{S} is then simply:
\begin{equation} \label{eq:samplesc}
    \textbf{S} = \textbf{B}_{n}.
\end{equation}
For notation purposes, this iterative bundling (Equation \ref{eq:samplesa}-\ref{eq:samplesc}) will be written in short as:
\begin{equation} \label{eq:samples}
    \textbf{S} = \bigoplus_{j=1}^{n} \textbf{v}_{x[j]}
\end{equation}
which is binarized into the HV $\textbf{s} = [\textbf{S}]$ with the majority rule (Equation \ref{eq:majority}). \\

\noindent Additionally, when the input data is encoded as $n$-grams, permutations can be used for the encoding (\cite{Rahimi2016a}):
\begin{equation*}
    \textbf{B}_{0} = \{0\}^D
\end{equation*}
\begin{equation} \label{eq:ngram}
    \textbf{B}_{j} = \rho(\textbf{B}_{j-1}) \oplus \textbf{s}_{j}
\end{equation}
\begin{equation*}
    \textbf{s}_{n-gram} = [\textbf{S}_{n-gram}] = [\textbf{B}_{n}]
\end{equation*}
where $j = 1...n$, $\textbf{s}_{n-gram}$ is the encoded, binarized HV of the $n$-gram of samples, and $\textbf{s}_{j}$ is a sample HV calculated with Equation \ref{eq:samples} followed by binarization (Equation \ref{eq:majority}). \\
    
\noindent \textbf{(3) Initial prototype construction.} Sample HVs $\textbf{s}_{i}$ belonging to the same class $l$ are bundled to form a class bundle $\textbf{C}_{l}$ representing the considered class:
\begin{equation}
    \textbf{C}_{l} = \bigoplus_{i=1}^{m} \{\textbf{s}_{i} | y_{i} = l\}
\end{equation}
with $y_{i}$ the $i$th sample's class. After binarization of the class bundle $\textbf{C}_{l}$, the $l$th class prototype $\textbf{c}_{l} = [\textbf{C}_{l}]$ is obtained and stored in the Associative Memory (AM). This is repeated for each class present in the dataset. \\

\noindent \textbf{(4) Training.} The HDC classifier predicts the class for all training samples by calculating the similarity between the training sample's HV $\textbf{s}_{i}$ and each class prototype $\textbf{c}_{k}$ stored in the AM. The predicted class $\hat{y}_{i}$ of the input sample is the class with the highest similarity to the input sample's HV:
\begin{equation} \label{eq:prediction}
    \hat{y}_{i} = \argmax_{k} s(\textbf{s}_{i},\textbf{c}_{k})
\end{equation} \\
If the predicted class is correct (i.e., $\hat{y}_{i} = y_{i} = l$), nothing happens. However, if the sample HV $\textbf{s}_{i}$ is wrongly classified (i.e., $\hat{y}_{i} \neq y_{i}$), it is bundled again in the class bundle of the correct class $\textbf{C}_{l}$ and bundled out of the class bundle of the wrong class $\textbf{C}_{\hat{l}}$:
\begin{equation} \label{eq:update1}
    \textbf{C}_{l} = \textbf{C}_{l} \oplus \textbf{s}_{i}
\end{equation}
\begin{equation} \label{eq:update2}
    \textbf{C}_{\hat{l}} = \textbf{C}_{\hat{l}} \ominus \textbf{s}_{i}
\end{equation}
with $\ominus$: $\mathcal{B} \times \mathcal{H} \to \mathcal{B}$: $(\textbf{B},\textbf{v}) \to \textbf{B} - \textbf{v}$ the bundling out operation which is element-wise subtraction. As such, the class prototypes are adjusted to better classify wrongly classified samples. \\
The training procedure is performed iteratively until either a predefined accuracy on the training set is reached or either a predefined number of iterations is performed. After each iteration, the updated class bundles are binarized into updated class prototypes to be used in the next iteration or finally, in the inference phase. \\

\noindent \textbf{(5) Inference.} The $i$th test sample is encoded in a query HV $\textbf{q}_{i}$ following the same encoding procedure as for the training samples. The predicted class is obtained similarly as during the training procedure with Equation \ref{eq:prediction} where $\textbf{s}_{i} = \textbf{q}_{i}$, i.e., the predicted class label of the test sample is the class with the highest similarity to the test samples's HV $\textbf{q}_{i}$.

\section{Previously proposed adjustments to HDC}

Already several suggestions have been made to improve the initial prototype construction and the training procedure in the HDC framework. For instance, \textcite{Rahimi2016a} only add a sample vector to the class bundle if the similarity between the sample vector and class vector is smaller than 0.9 such that a sample vector is not added to the class bundle if the class vector is already highly similar to the sample vector. Consequently, no redundant information is added and the initial prototypes are assumed to be better, reducing the training time afterwards. \textcite{Imani2019a} propose to perform the training procedure with an adaptive learning rate that depends (1) on the average error rate over the last few training iterations (= iteration-dependent learning), (2) on the difference in similarity between query and wrong class on the one hand and query and right class on the other hand (= data-dependent learning) or (3) a combination of both (= hybrid learning). During the training process explained by \textcite{Hernandez2021a}, sample vectors are added to or subtracted from class bundles with a weight depending on the similarity of that sample vector to the considered class vector. As such, a sample with high similarity will be added with smaller weight than a sample with low similarity, since the sample is already highly represented by the class vector in case of high similarity and thus redundant information in the class prototype is limited. \\

\noindent In addition, more complex methods to improve HDC have been introduced in literature such as applying manifold learning for unsupervised non-linear dimensionality reduction to project the original data to a smaller dimension before applying HD encoding (\cite{Zou2021a}). \textcite{Hsiao2021} use a learnable projection to train the HDC in a similar way as a binary neural network. The trained binary weights are then transformed into learned IM and CIM. A different way of \textit{linear mapping} is proposed by \textcite{Basaklar2021} where a variable number of bits are flipped between levels to emphasize the distinct levels, instead of flipping a uniform number of bits. \textcite{Chuang2020} use a confidence metric to decide whether binary HDC is suitable for a specific sample, i.e., samples predicted with low confidence in binary HDC will be predicted with non-binary HDC to improve the classification performance. On the other hand, \textcite{Duan2022a} map the HDC framework to an equivalent binary neural network (BNN) which is trained to minimize the training loss to increase the confidence in predictions. These methods in general achieve higher accuracies, but as they are more complex, they also have a large computational overhead. \\

\noindent In contrast, the focus of this article is on a simple, yet effective extension of the binary HDC training procedure that consistently improves the baseline performance, but without additional complexity. Its basic idea is to not only take into account wrong predictions when updating the class bundles, but also samples that are correctly classified but for which the class with the second highest similarity is only slightly less similar to the sample than the correct class. This simple extension of the HDC training procedure is explained in the next section.

\section{Our proposal: confidence-based training procedure}

As indicated by \textcite{Duan2022a}, for a correctly classified sample the similarity to the class with the second highest similarity could only be slightly lower than the similarity to the class with the highest similarity. In such cases, the HDC classifier is less confident about its prediction for that specific sample. Knowing this, a confidence metric $c_{i}$ is introduced by \textcite{Chuang2020} as the difference between similarity of the sample vector $\textbf{s}_{i}$ to the class vector with the highest similarity and the similarity to the class vector with the second highest similarity:
\begin{equation} \label{eq:confidence}
    c_{i} = \max_{k} s(\textbf{s}_{i},\textbf{c}_{k}) - \max_{k \neq l} s(\textbf{s}_{i},\textbf{c}_{k}).
\end{equation}
The confidence metric reflects with how much certainty the HDC model classifies a specific sample, i.e., if the confidence is low, the HDC model is less certain about its prediction of a specific sample. \\

\noindent While \textcite{Chuang2020} use the confidence metric only to measure certainty of classification on the test set, our proposal is to use this confidence metric in a simple way to extend the training procedure in HDC resulting in a more accurate binary HDC model. In all the basic HDC training frameworks, class bundles are only updated in case of a wrong prediction. However, in our proposal a training sample's HV $\textbf{s}_{i}$ that is correctly classified (i.e., $\hat{y}_{i} = y_{i} = l$) but with a confidence smaller than a threshold $\alpha$ (i.e., $c_{i} < \alpha$), is also added again to the class bundle of the correct class $\textbf{C}_{l}$ and bundled out of the class bundle of the class with second highest similarity $\textbf{C}_{\hat{l}'}$ with $\hat{l}' = \argmax_{k \neq l} s(\textbf{s}_{i},\textbf{c}_{k})$:
\begin{equation} \label{eq:update3}
    \textbf{C}_{l} = \textbf{C}_{l} \oplus \textbf{s}_{i}
\end{equation}
\begin{equation} \label{eq:update4}
    \textbf{C}_{\hat{l}'} = \textbf{C}_{\hat{l}'} \ominus \textbf{s}_{i}
\end{equation}
Note that if $\alpha = 0$, no correctly classified samples are used in the training procedure since the confidence is non-negative. As a consequence, this threshold setting corresponds to the baseline HDC classifier. Updating the class bundles with samples correctly classified with low confidence could be seen as pulling the wrong class further away from the considered sample and pushing the right class closer to the sample. As such, the main idea of this updating procedure has some analogies to prototype learning (\cite{Chang2006, Ji2021}), prototype alignment (\cite{Hersche2022}), distance metric learning (\cite{Weinberger2009,Kulis2012}), linear discriminant analysis (\cite{Weinberger2009}) and support vector machines (\cite{Weinberger2009}) where the goal is to minimize the distance between samples from the same class while maximizing the distance between samples from different classes. \\

\noindent With this proposal of a simple extended training procedure, an improved classification performance is expected since the class prototypes are not only adjusted to better classify wrongly classified samples (Equation \ref{eq:update1}-\ref{eq:update2}), but also samples that are correctly classified with low confidence (Equation \ref{eq:update3}-\ref{eq:update4}).

\section{Experiments}

Four datasets, that are publicly available and commonly or previously used in other HDC-related research, are selected to test the proposed extended training procedure (a more detailed summary of the datasets can be found in Table \ref{tab:datasets}): \\
\textbf{(1) UCIHAR dataset} (\cite{UCIHAR2013,UCI2019}). To obtain this dataset, 30 subjects performed six activities (walking, walking upstairs, walking downstairs, sitting, standing and laying) during which the acceleration and velocity were recorded with the accelerometer and gyroscope of a smartphone attached to the chest of the subjects. \\
\textbf{(2) Cardiotocography (CTG) dataset} (\cite{UCI2019}). This dataset consists of features for 2,126 fetal cardiotocograms that are classified with respect to one of three fetal states (normal, suspect or pathologic). \\
\textbf{(3) ISOLET dataset} (\cite{UCI2019}). This dataset includes features extracted from speech signals that were collected for 150 subjects speaking each letter of the alphabet twice. \\
\textbf{(4) Hand gesture (HAND) dataset} (\cite{Rahimi2016a}). EMG signals with four channels of five subjects are included in this dataset. During the recording, the subjects performed four hand gestures (closed hand, open hand, 2-finger pinch and point index) ten times for three seconds each. Between each hand gesture contraction, a period of three seconds in the rest position is included which serves as the fifth class. \\

\noindent To quantize the feature / signal values of all datasets, a step size is chosen such that they are quantized into 21 quantization levels. As such, a step size of 0.1 is chosen for UCIHAR because values are between -1 and 1, a step size of 5 for CTG's values ranging from 0 to 100, a step size of 10 for ISOLET since the values range from -100 to 100 and a step size of 1 for HAND because the signals can take amplitudes from 0 to 20. The samples of UCIHAR, CTG and ISOLET are encoded following Equation \ref{eq:samples} followed by binarization (Equation \ref{eq:majority}), whereas $4-grams$ (Equation \ref{eq:ngram}) are created for the encoding of HAND samples that are also encoded according to Equation \ref{eq:samples} with binarization (Equation \ref{eq:majority}) where the four channels are treated as four features. \\

\begin{table}[!hbt]
\centering
\caption{Summary of the four datasets used to test the proposed extended training procedure. The table includes the number of training samples, the number of test samples, the number of classes, the number of features or channels, the range of values of the features, the step size that is used to quantize the feature values and the number of vectors (i.e., quantization levels) that are stored in the CIM.}
\label{tab:datasets}
\begin{tabular}{@{}rrrrr@{}}
\hline
 & \textbf{UCIHAR} & \textbf{CTG} & \textbf{ISOLET} & \textbf{HAND} \\
\hline
\textbf{\# training samples}      & 7,352           & 1,701           & 6,238          & 526,396       \\
\textbf{\# test samples}          & 2,947           & 425             & 1,559          & 131,608       \\
\textbf{\# classes}               & 6               & 3               & 26             & 5             \\
\textbf{\# features / channels}   & 561             & 21              & 617            & 4             \\
\textbf{range of feature values}  & {[}-1,1{]}      & {[}0,100{]}     & {[}-100,100{]} & {[}0,20{]}    \\
\textbf{quantization step size}   & 0.1             & 5               & 10             & 1             \\
\textbf{\# vectors in CIM}        & 21              & 21              & 21             & 21            \\ \hline
\end{tabular}
\end{table}

\noindent The training procedure is performed iteratively for a maximum of 2500 iterations while saving the classifier with the best accuracy. After every 100 iterations, it is evaluated whether this best training accuracy exceeds 99\%. If this is the case, the training procedure is terminated and the classifier with the best accuracy is used in the inference phase. The HDC classifier is performed for 50 independent runs for UCIHAR, CTG and ISOLET, and for 10 independent runs for each subject of HAND (thus, also 50 independent runs in total) since it starts from random vectors to form the atomic vectors. \\

\begin{figure}[h]
\begin{subfigure}{.5\linewidth}
    \centering
    \includegraphics[scale=.3]{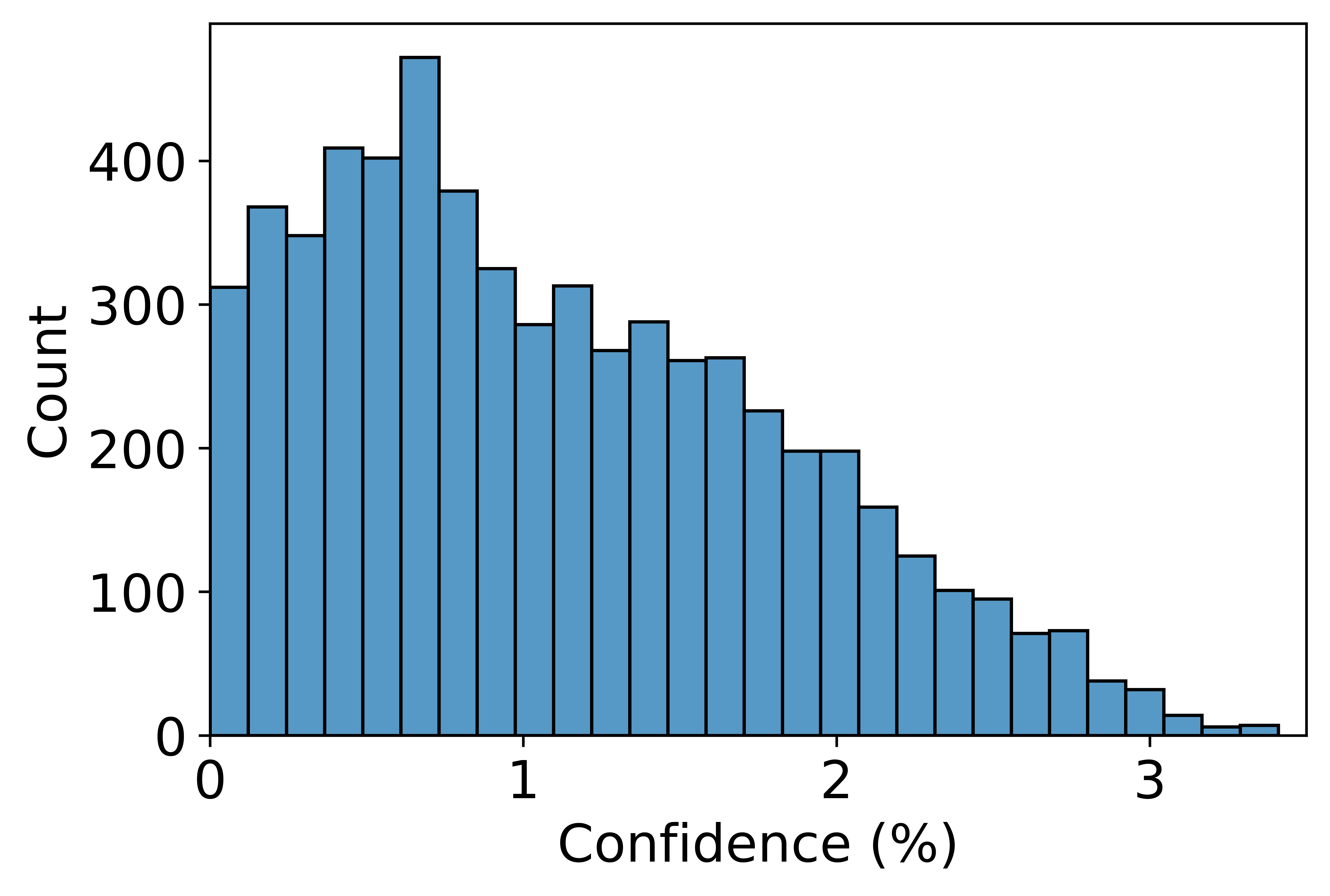}
    \caption{UCIHAR.}
    \label{fig:confUCIHAR}
\end{subfigure}
\begin{subfigure}{.5\linewidth}
    \centering
    \includegraphics[scale=.3]{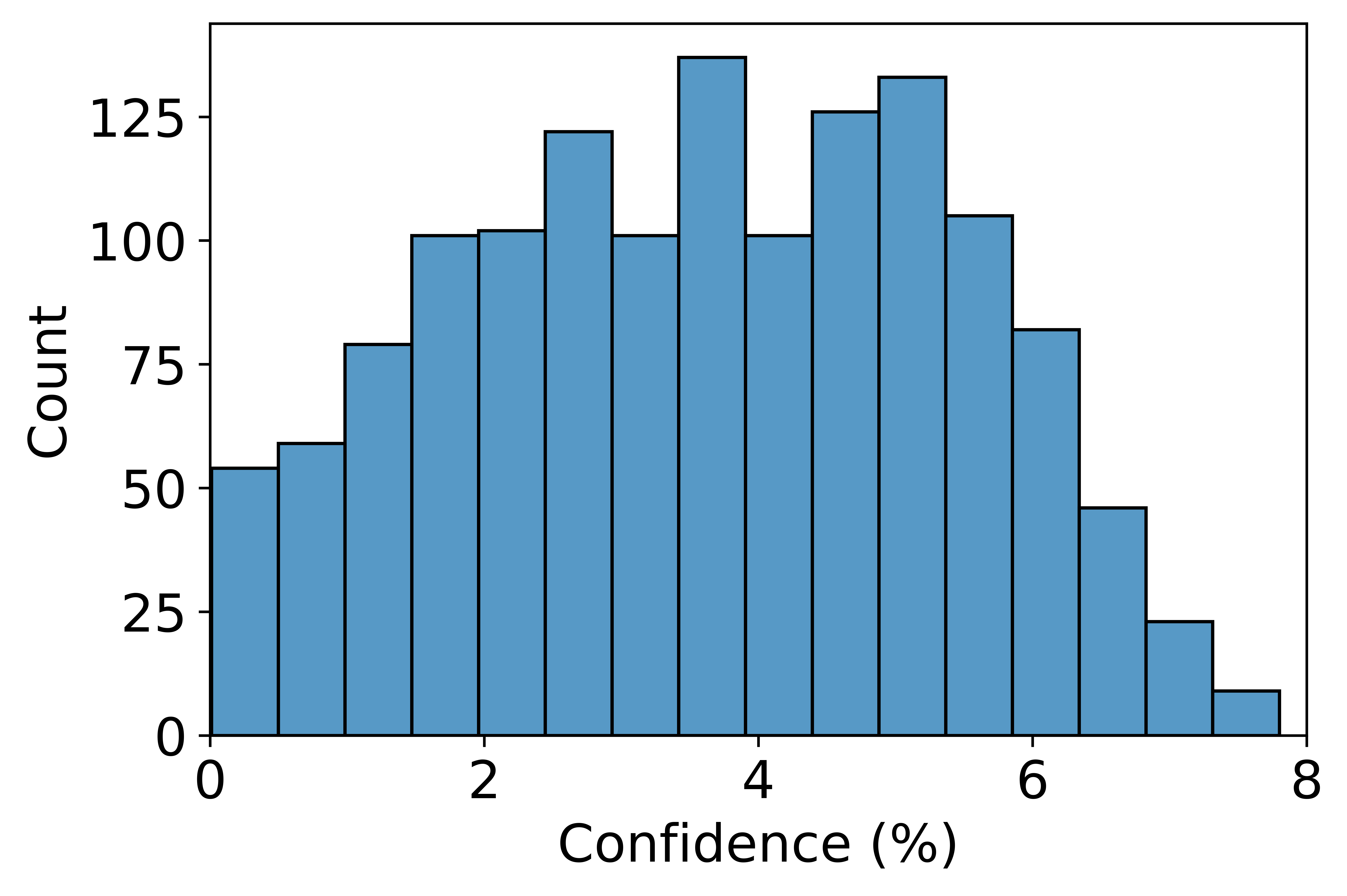}
    \caption{CTG.}
    \label{fig:confCTG}
\end{subfigure}
\vskip \baselineskip
\vspace{-10pt}
\begin{subfigure}{.5\linewidth}
    \centering
    \includegraphics[scale=.3]{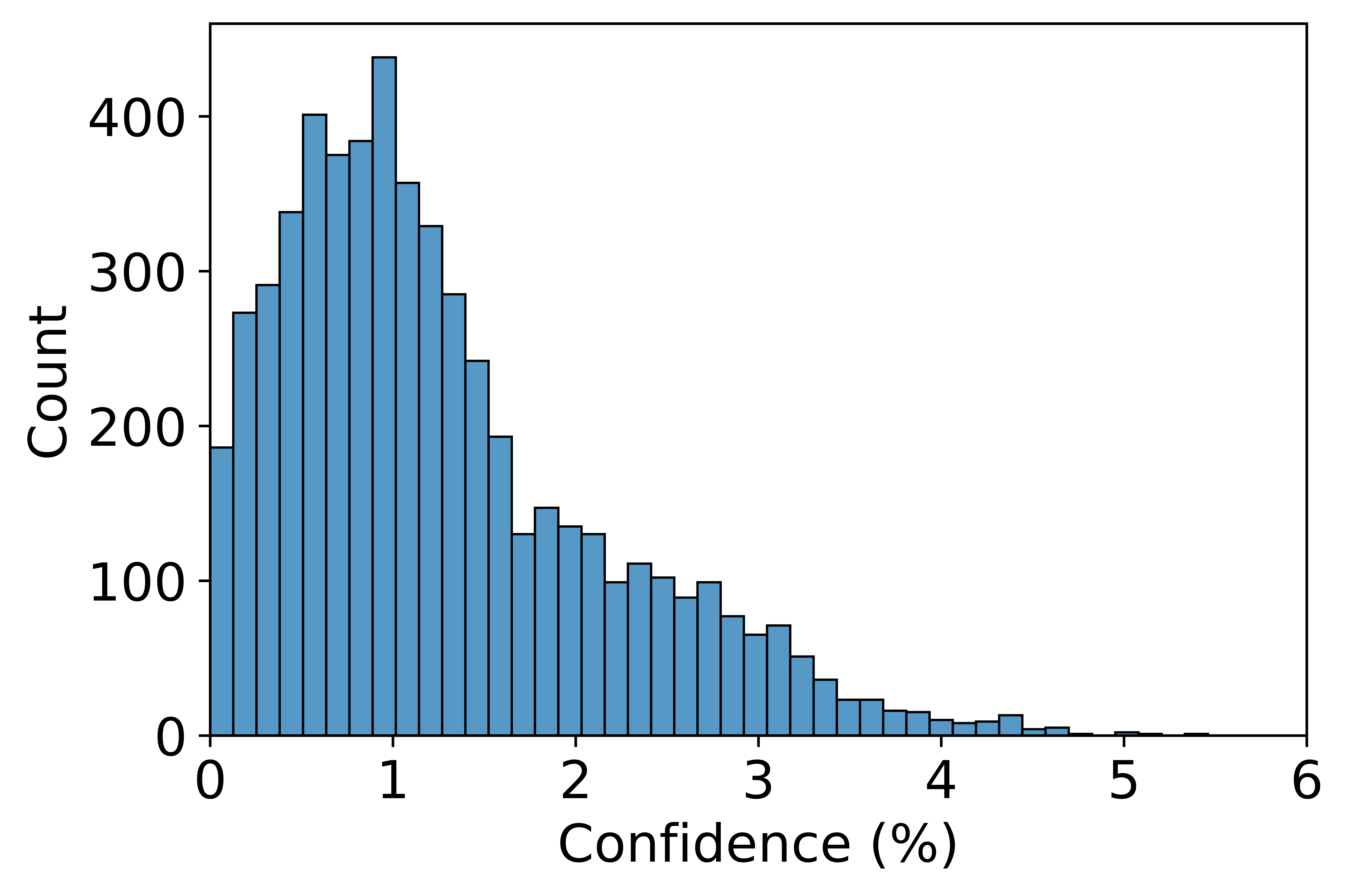}
    \caption{ISOLET.}
    \label{fig:confISOLET}
\end{subfigure}
\begin{subfigure}{.5\linewidth}
    \centering
    \includegraphics[scale=.3]{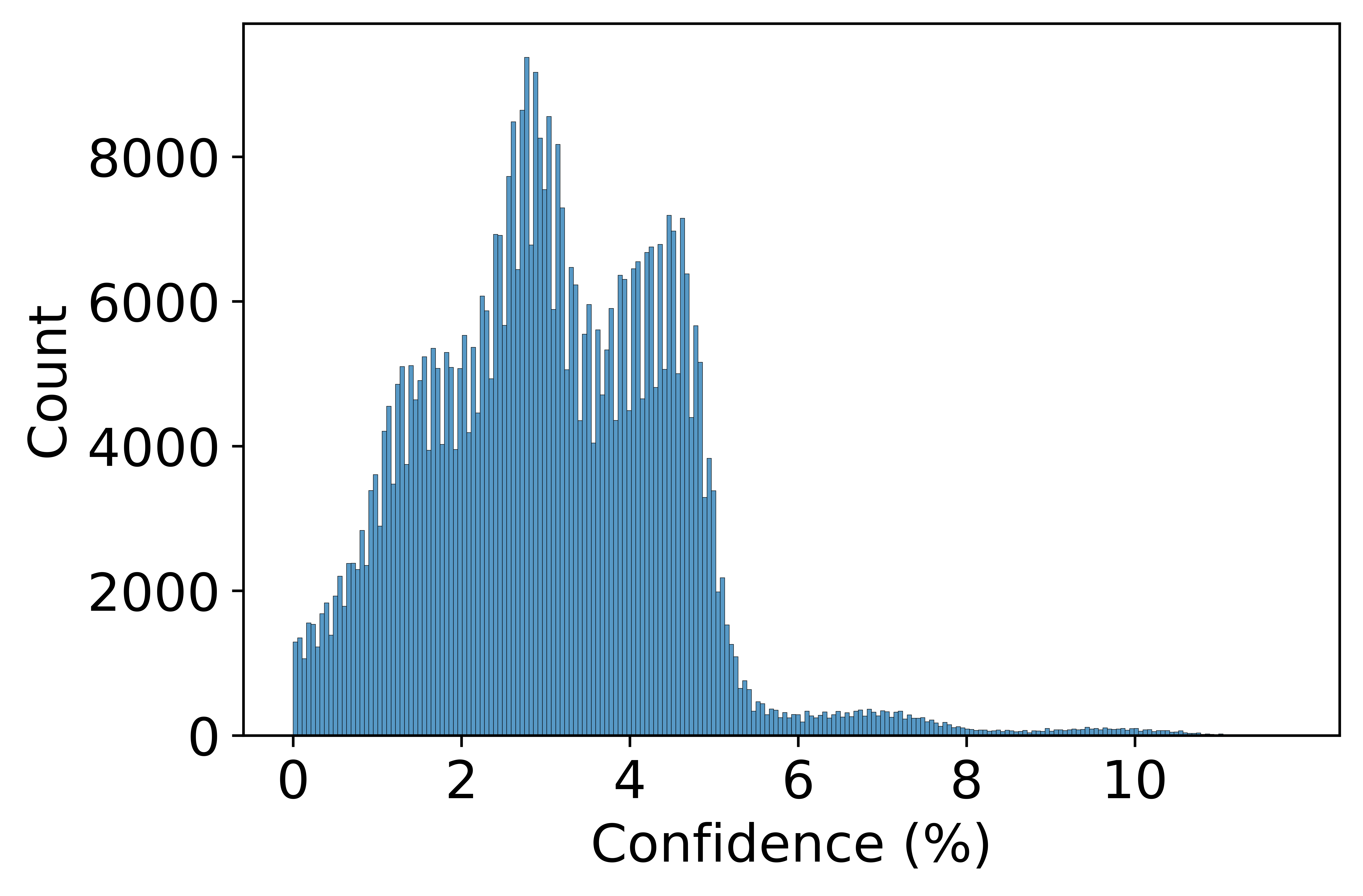}
    \caption{HAND.}
    \label{fig:confHAND}
\end{subfigure}
\caption{Distribution of the confidence values of all correctly classified training samples of (a) UCIHAR, (b) CTG, (c) ISOLET and (d) HAND after initial prototype construction.}
\label{fig:conf}
\end{figure}

\noindent For each dataset, the distribution of confidence values of all correctly classified training samples after initial prototype construction is investigated to decide the range of confidence thresholds to be tested (Figure \ref{fig:conf}). This figure illustrates that the confidence values are rather low for all datasets (i.e., $<$ 8\%). Hence, the choice for the thresholds to be tested are $\alpha = \{0.00;$ $0.25;$ $0.50;$ $0.75;$ $1.00;$ $1.25;$ $1.50\}$ for UCIHAR, $\alpha = \{0.00;$ $1.00;$ $2.00;$ $3.00;$ $4.00;$ $5.00;$ $6.00\}$ for CTG, $\alpha = \{0.00;$ $0.25;$ $0.50;$ $0.75;$ $1.00;$ $1.25;$ $1.50\}$ for ISOLET and $\alpha = \{0.00;$ $1.00;$ $2.00;$ $3.00;$ $4.00;$ $5.00\}$ for HAND, since these values include most correctly classified samples for the considered datasets. The performance of the HDC classifier for each confidence threshold setting is documented as the classification accuracy and error rate on the test set averaged over all performed independent runs.

\section{Results and discussion}

The accuracies on the train and test set and the error rate on the test set averaged over all independent runs for each of the chosen confidence threshold settings and each of the four datasets are given in Table \ref{tab:results}. The introduction of the confidence metric in the training procedure improves the baseline classification accuracy ($\alpha = 0.00$) for each dataset for all tested confidence thresholds. The mean test accuracy for all datasets increases with increasing $\alpha$ up to the point where maximal performance is reached after which the mean test accuracy decreases for larger $\alpha$. \\
\textbf{(1) UCIHAR.} The obtained baseline accuracy of 92.75\% is improved by 1.58\% with $\alpha = 0.75$ reaching an accuracy of 94.33\%. (A 21.79\% relative decrease in error rate from 7.25\% for the baseline to 5.67\%.) \\
\noindent \textbf{(2) CTG.} The obtained baseline accuracy of 73.65\% is improved by 13.24\% with $\alpha = 4.00$ reaching an accuracy of 86.89\%. (A 50.25\% relative decrease in error rate from 26.35\% for the baseline to 13.11\%.) \\
\textbf{(3) ISOLET.} The obtained baseline accuracy of 92.12\% is improved by 2.18\% with $\alpha = 1.00$ reaching an accuracy of 94.30\%. (A 27.66\% relative decrease in error rate from 7.88\% for the baseline to 5.70\%.) \\
\textbf{(4) HAND.} The obtained baseline accuracy of 95.38\% is improved by 0.93\% with $\alpha = 5.00$ reaching an accuracy of 96.31\%. (A 20.13\% relative decrease in error rate from 4.62\% for the baseline to 3.69\%.) \\

\begin{table}[!ht]
\caption{Averaged accuracy (\%) on the train and test set and averaged error rate (\%) on test set over 50 independent runs (for UCIHAR, CTG and ISOLET) and 10 independent runs (for each subject of HAND, thus 50 in total) of the four datasets for the tested settings of the confidence threshold $\alpha$. Data are \textit{mean} ($\pm$ \textit{standard deviation}), in \textbf{bold} are the test accuracies (and error rates) that are higher (and lower) than the baseline ($\alpha = 0.00$) and \underline{underlined} is the best test accuracy and error rate.}
\label{tab:results}
\begin{subtable}[h]{\textwidth}
    \caption{UCIHAR.}
    \centering
    \begin{tabular}{@{}cccc@{}}
    \hline
    \textbf{$\alpha$} & \textbf{Train accuracy} & \textbf{Test accuracy} & \textbf{Test error rate} \\
    \hline
    0.00 & 99.43 ($\pm$ 0.24) & 92.75 ($\pm$ 0.51) & 7.25 \\
    0.25 & 99.11 ($\pm$ 0.20) & \textbf{93.94} ($\pm$ 0.50) & \textbf{6.06} \\
    0.50 & 98.96 ($\pm$ 0.51) & \textbf{94.03} ($\pm$ 0.53) & \textbf{5.97} \\
    0.75 & 98.62 ($\pm$ 0.61) & \underline{\textbf{94.33}} ($\pm$ 0.49) & \underline{\textbf{5.67}} \\
    1.00 & 98.32 ($\pm$ 0.99) & \textbf{94.25} ($\pm$ 0.70) & \textbf{5.75} \\
    1.25 & 97.13 ($\pm$ 1.15) & \textbf{94.16} ($\pm$ 0.62) & \textbf{5.84} \\
    1.50 & 95.99 ($\pm$ 1.19) & \textbf{93.45} ($\pm$ 0.90) & \textbf{6.55} \\ \hline
    \end{tabular}
    \label{tab:resultsUCIHAR}
\end{subtable}
\vskip \baselineskip
\vspace{-10pt}
\begin{subtable}[h]{\textwidth}
    \caption{CTG.}
   \centering
   \begin{tabular}{@{}cccc@{}}
   \hline
   \textbf{$\alpha$} & \textbf{Train accuracy} & \textbf{Test accuracy} & \textbf{Test error rate} \\
   \hline
   0.00 & 99.57 ($\pm$ 0.21) & 73.65 ($\pm$ 2.48) & 26.35 \\
   1.00 & 98.91 ($\pm$ 0.17) & \textbf{77.52} ($\pm$ 2.53) & \textbf{22.48} \\
   2.00 & 98.05 ($\pm$ 0.13) & \textbf{80.44} ($\pm$ 1.79) & \textbf{19.56} \\
   3.00 & 96.39 ($\pm$ 0.17) & \textbf{85.14} ($\pm$ 1.28) & \textbf{14.86} \\
   4.00 & 94.47 ($\pm$ 0.29) & \underline{\textbf{86.89}} ($\pm$ 1.04) & \underline{\textbf{13.11}} \\
   5.00 & 93.19 ($\pm$ 0.27) & \textbf{86.19} ($\pm$ 0.94) & \textbf{13.81} \\
   6.00 & 92.47 ($\pm$ 0.14) & \textbf{84.97} ($\pm$ 1.01) & \textbf{15.03} \\ \hline
   \end{tabular}
   \label{tab:resultsCTG}
\end{subtable}
\vskip \baselineskip
\vspace{-10pt}
\begin{subtable}[h]{\textwidth}
   \caption{ISOLET.}
   \centering
   \begin{tabular}{@{}cccc@{}}
   \hline
   \textbf{$\alpha$} & \textbf{Train accuracy} & \textbf{Test accuracy} & \textbf{Test error rate} \\
   \hline
   0.00 & 100.00 ($\pm$ 0.00) & 92.12 ($\pm$ 0.47) & 7.88 \\
   0.25 & 100.00 ($\pm$ 0.00) & \textbf{93.19} ($\pm$ 0.42) & \textbf{6.81} \\
   0.50 & 99.99 ($\pm$ 0.03)  & \textbf{93.80} ($\pm$ 0.45) & \textbf{6.20} \\
   0.75 & 99.83 ($\pm$ 0.27)  & \textbf{94.13} ($\pm$ 0.55) & \textbf{5.87} \\
   1.00 & 99.81 ($\pm$ 0.22)  & \underline{\textbf{94.30}} ($\pm$ 0.73) & \underline{\textbf{5.70}} \\
   1.25 & 99.70 ($\pm$ 0.28)  & \textbf{93.98} ($\pm$ 0.75) & \textbf{6.02} \\
   1.50 & 99.47 ($\pm$ 0.25)  & \textbf{93.68} ($\pm$ 0.76) & \textbf{6.32} \\ \hline
   \end{tabular}
   \label{tab:resultsISOLET}
\end{subtable}
\vskip \baselineskip
\vspace{-10pt}
\begin{subtable}[h]{\textwidth}
    \caption{HAND.}
    \centering
    \begin{tabular}{@{}cccc@{}}
    \hline
    \textbf{$\alpha$} & \textbf{Train accuracy} & \textbf{Test accuracy} & \textbf{Test error rate} \\
    \hline
    0.00 & 97.05 ($\pm$ 1.84) & 95.38 ($\pm$ 1.82) & 4.62 \\
    1.00 & 96.94 ($\pm$ 1.92) & \textbf{95.77} ($\pm$ 1.83) & \textbf{4.23} \\
    2.00 & 96.73 ($\pm$ 1.85) & \textbf{95.60} ($\pm$ 2.25) & \textbf{4.40} \\
    3.00 & 96.45 ($\pm$ 2.01) & \textbf{95.81} ($\pm$ 2.03) & \textbf{4.19} \\
    4.00 & 96.07 ($\pm$ 2.20) & \textbf{96.11} ($\pm$ 1.65) & \textbf{3.89} \\
    5.00 & 95.83 ($\pm$ 2.26) & \underline{\textbf{96.31}} ($\pm$ 1.34) & \underline{\textbf{3.69}} \\ \hline
    \end{tabular}
    \label{tab:resultsHAND}
\end{subtable}
\end{table}

\begin{figure}[h]
\begin{subfigure}{.24\linewidth}
    \centering
    \includegraphics[scale=.24]{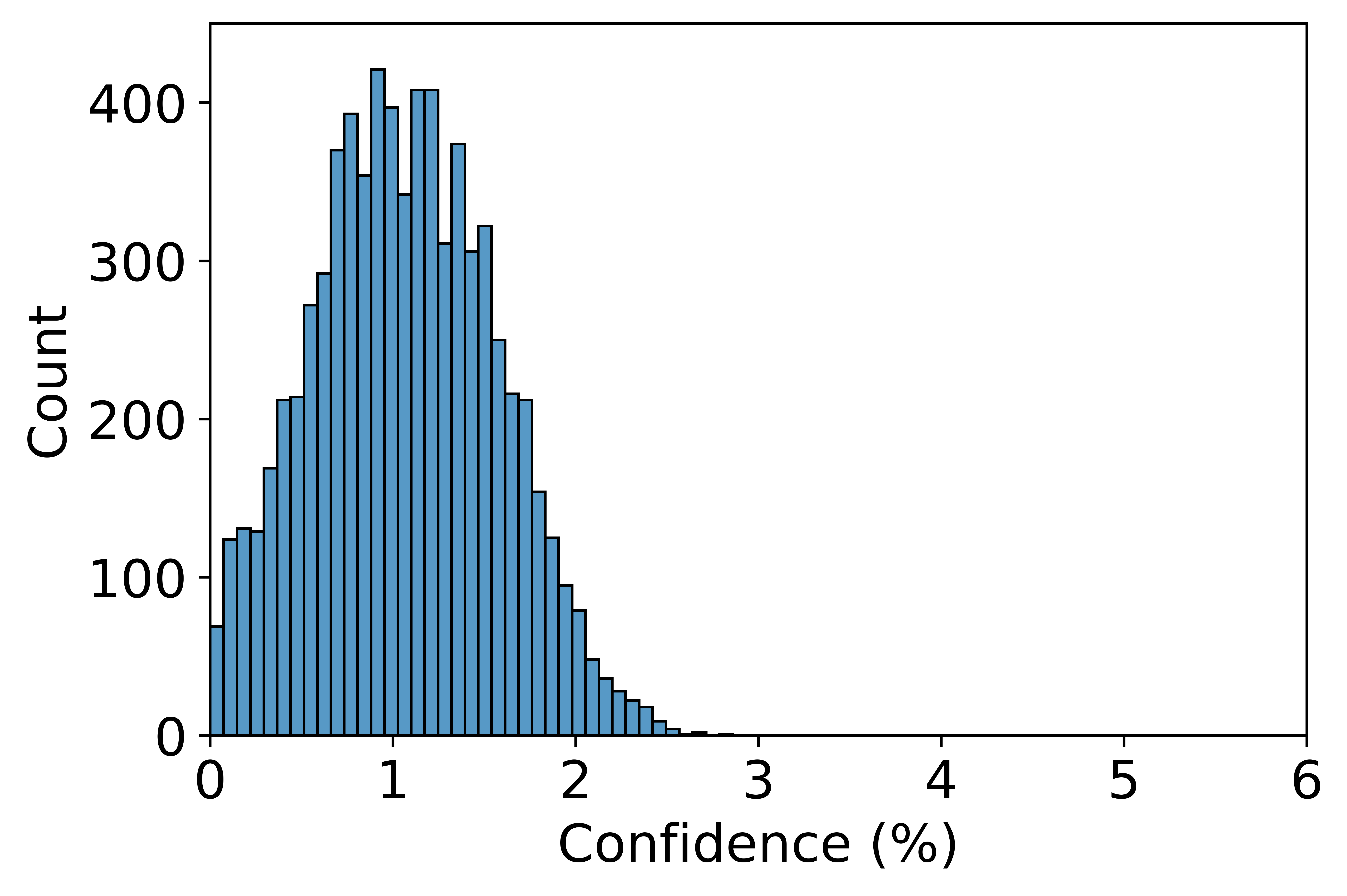}
    \caption{UCIHAR \\ $\alpha = 0.00$.}
    \label{fig:confUCIHAR0.00}
\end{subfigure}
\begin{subfigure}{.24\linewidth}
    \centering
    \includegraphics[scale=.24]{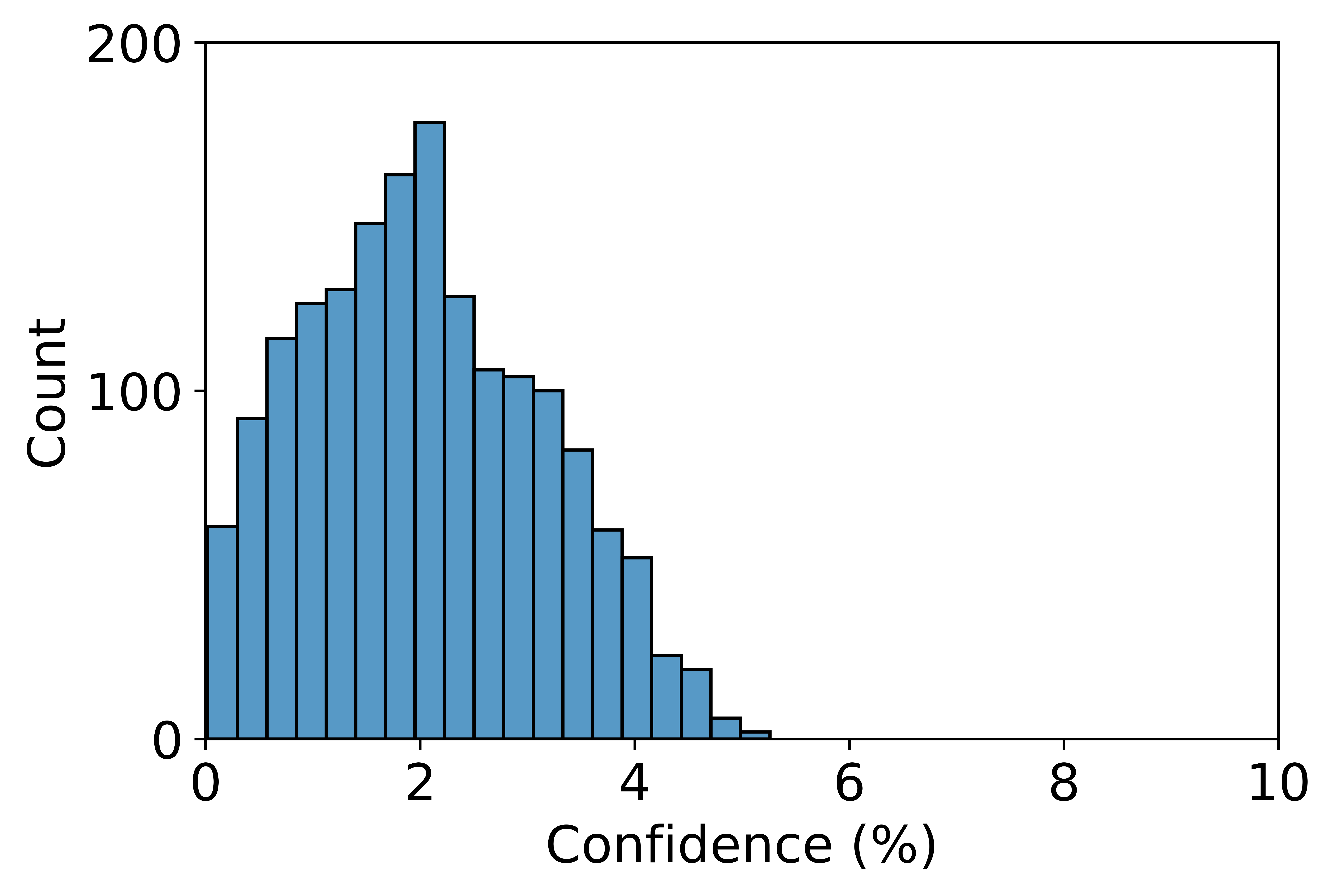}
    \caption{CTG \\ $\alpha = 0.00$.}
    \label{fig:confCTG0.00}
\end{subfigure}
\begin{subfigure}{.24\linewidth}
    \centering
    \includegraphics[scale=.24]{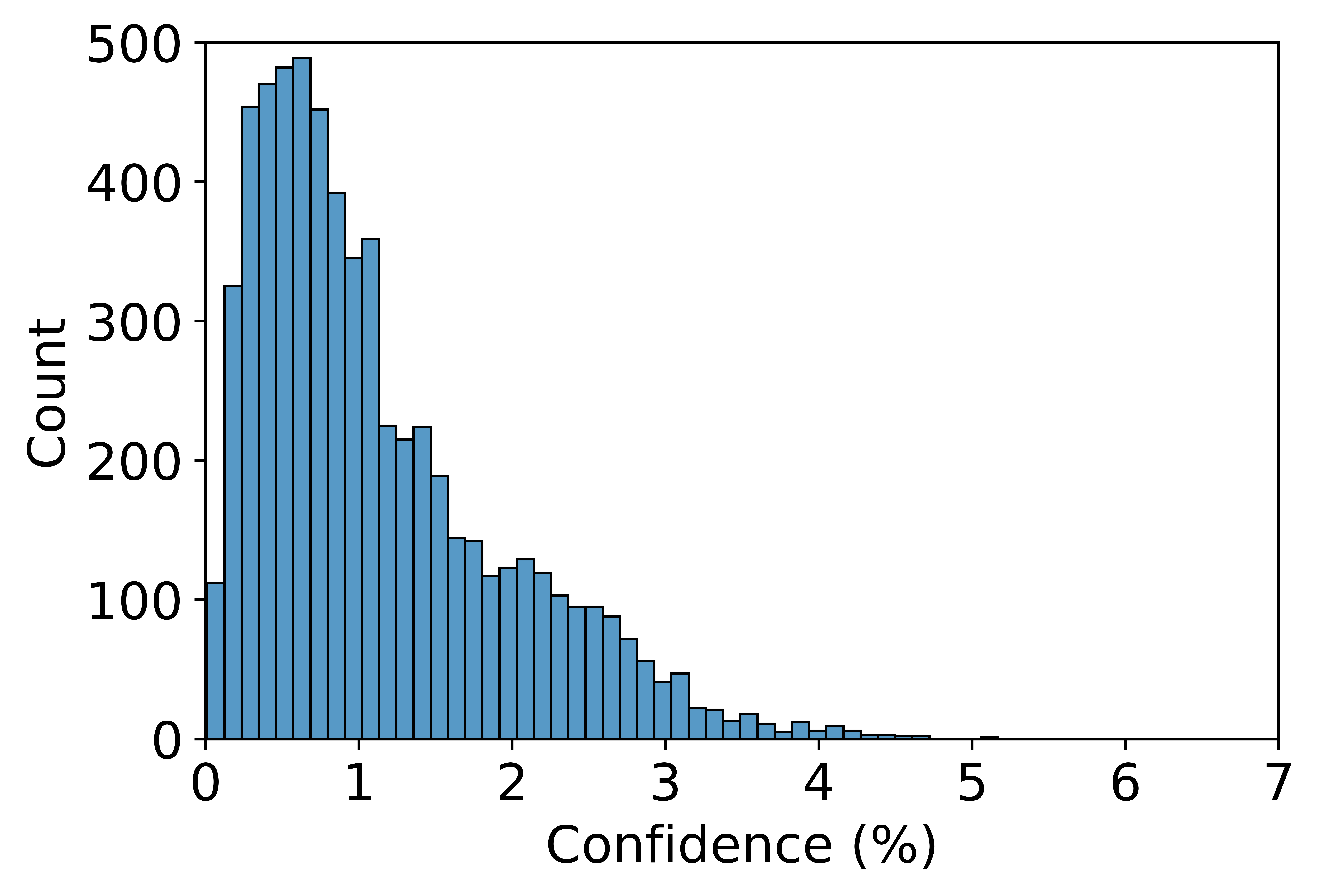}
    \caption{ISOLET \\ $\alpha = 0.00$.}
    \label{fig:confISOLET0.00}
\end{subfigure}
\begin{subfigure}{.24\linewidth}
    \centering
    \includegraphics[scale=.24]{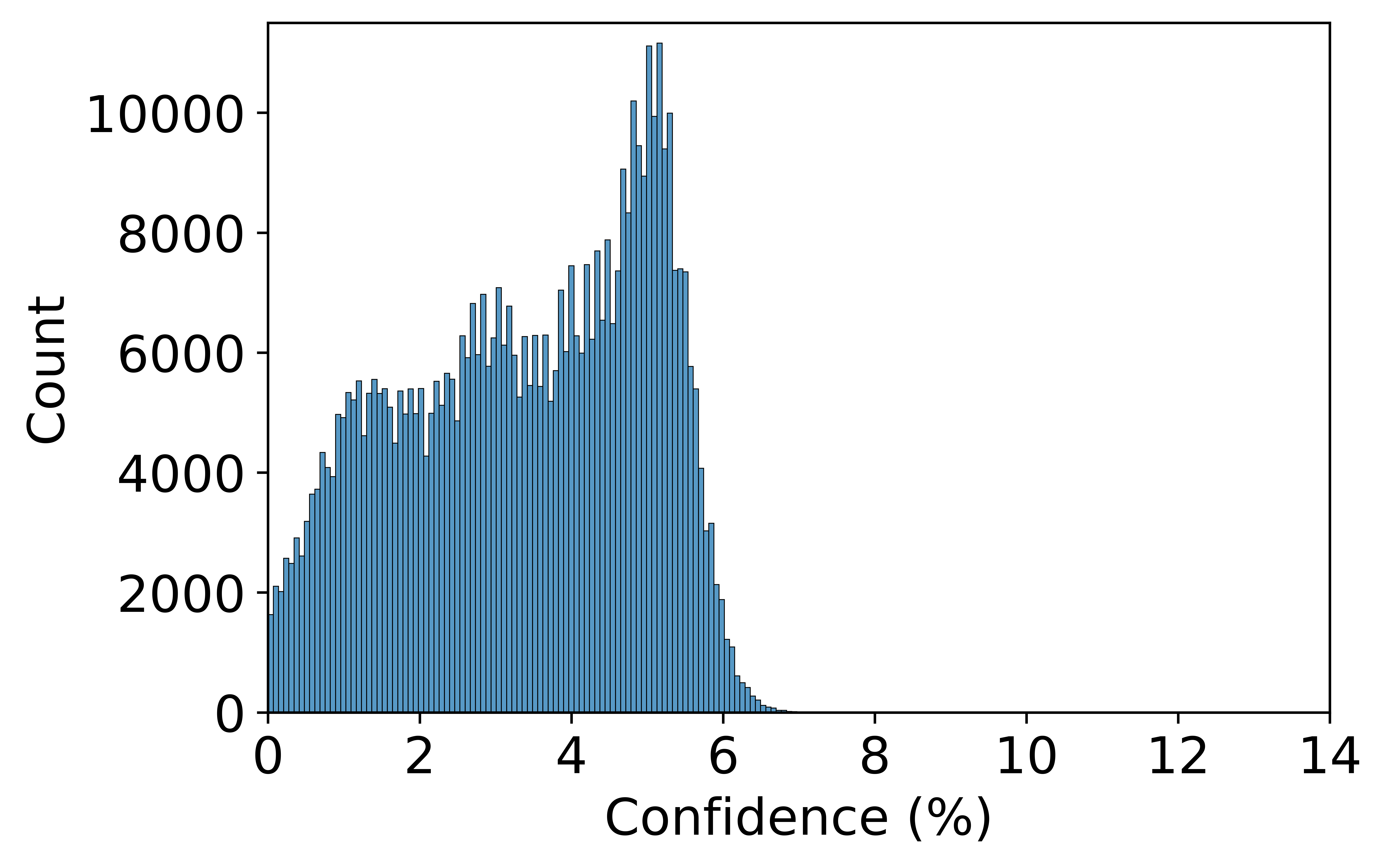}
    \caption{HAND \\ $\alpha = 0.00$.}
    \label{fig:confHAND0.00}
\end{subfigure}
\vskip \baselineskip
\vspace{-10pt}
\begin{subfigure}{.24\linewidth}
    \centering
    \includegraphics[scale=.24]{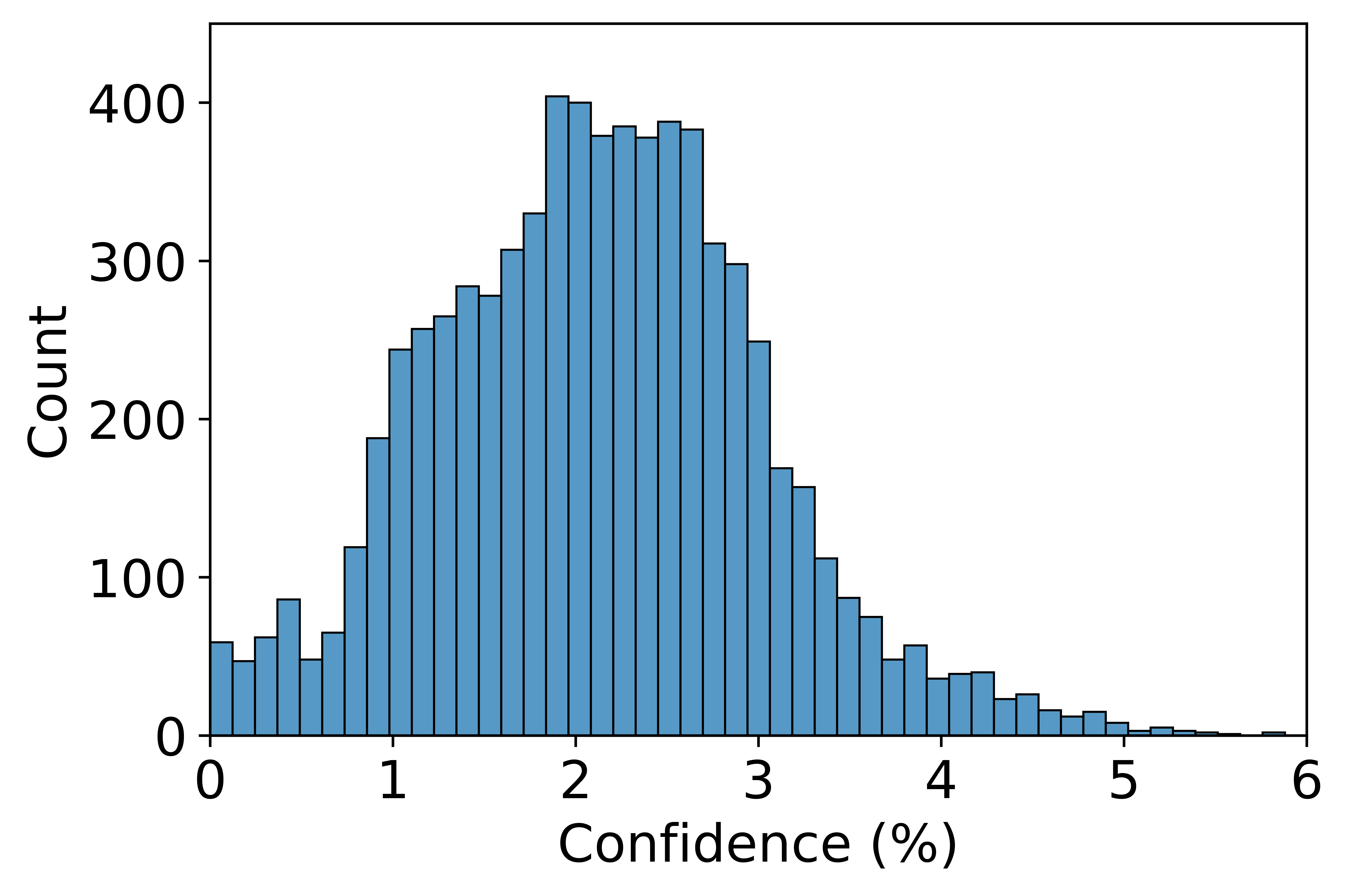}
    \caption{UCIHAR \\ $\alpha = 0.75$.}
    \label{fig:confUCIHAR0.75}
\end{subfigure}
\begin{subfigure}{.24\linewidth}
    \centering
    \includegraphics[scale=.24]{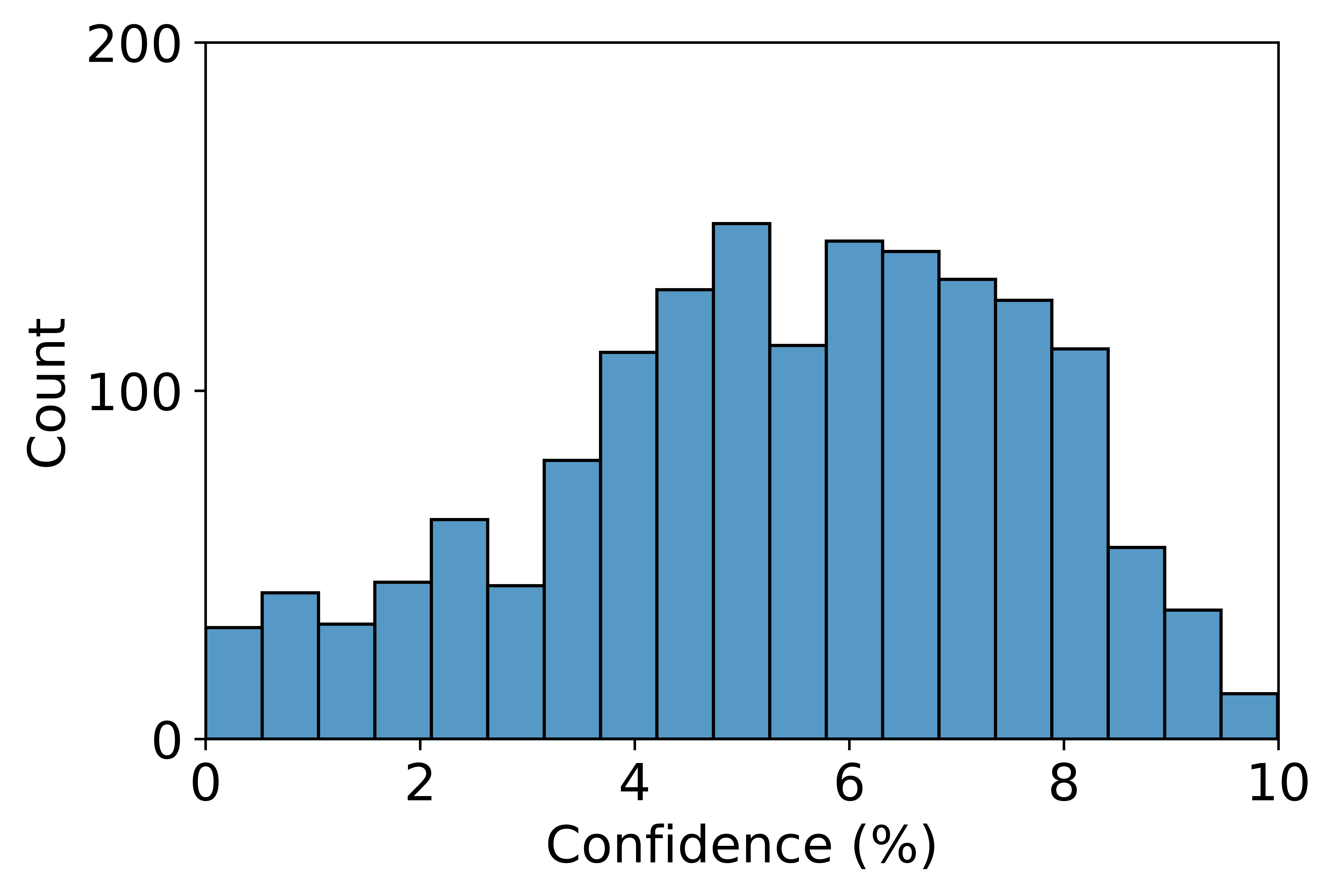}
    \caption{CTG \\ $\alpha = 4.00$.}
    \label{fig:confCTG4.00}
\end{subfigure}
\begin{subfigure}{.24\linewidth}
    \centering
    \includegraphics[scale=.24]{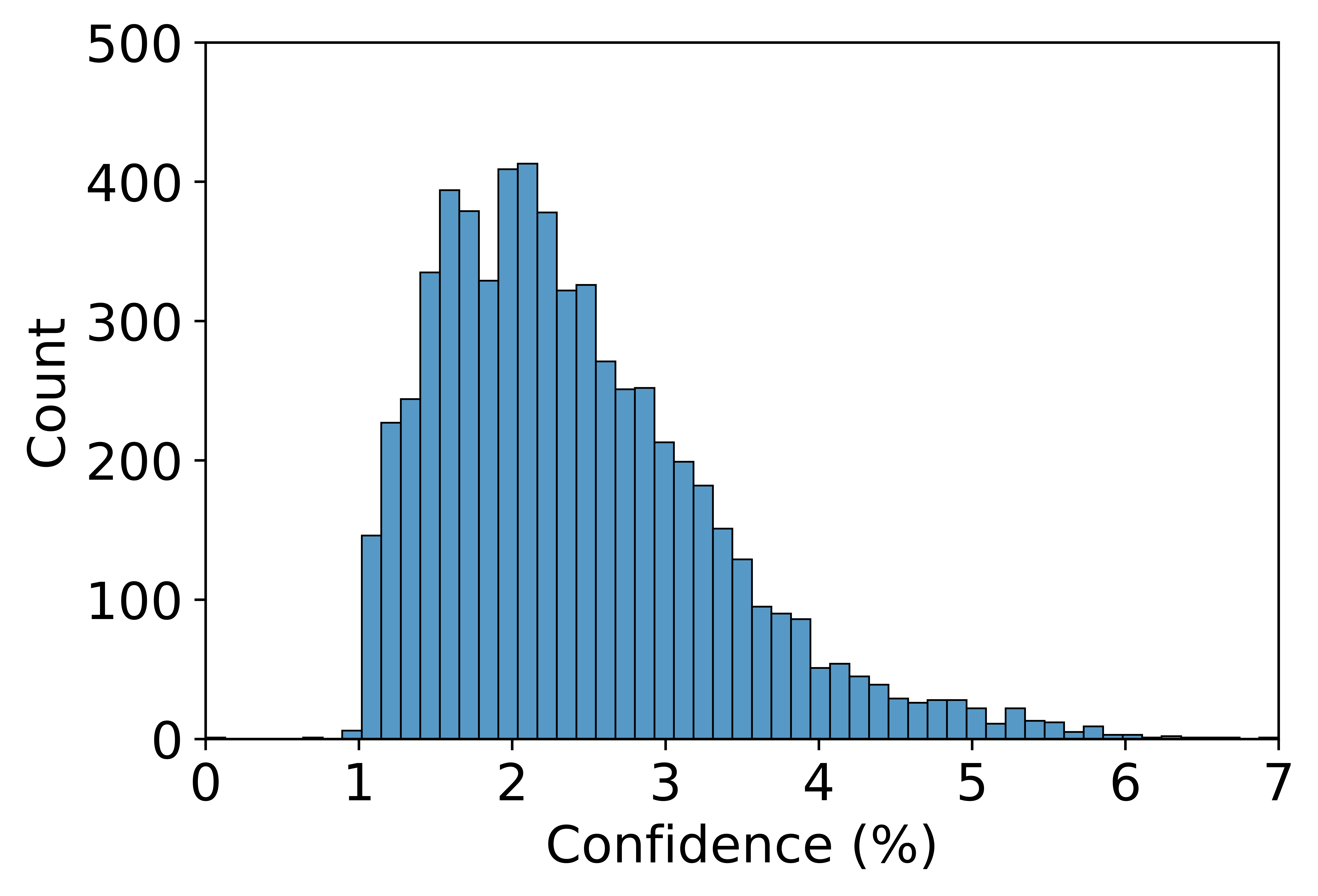}
    \caption{ISOLET \\ $\alpha = 1.00$.}
    \label{fig:confISOLET1.00}
\end{subfigure}
\begin{subfigure}{.24\linewidth}
    \centering
    \includegraphics[scale=.24]{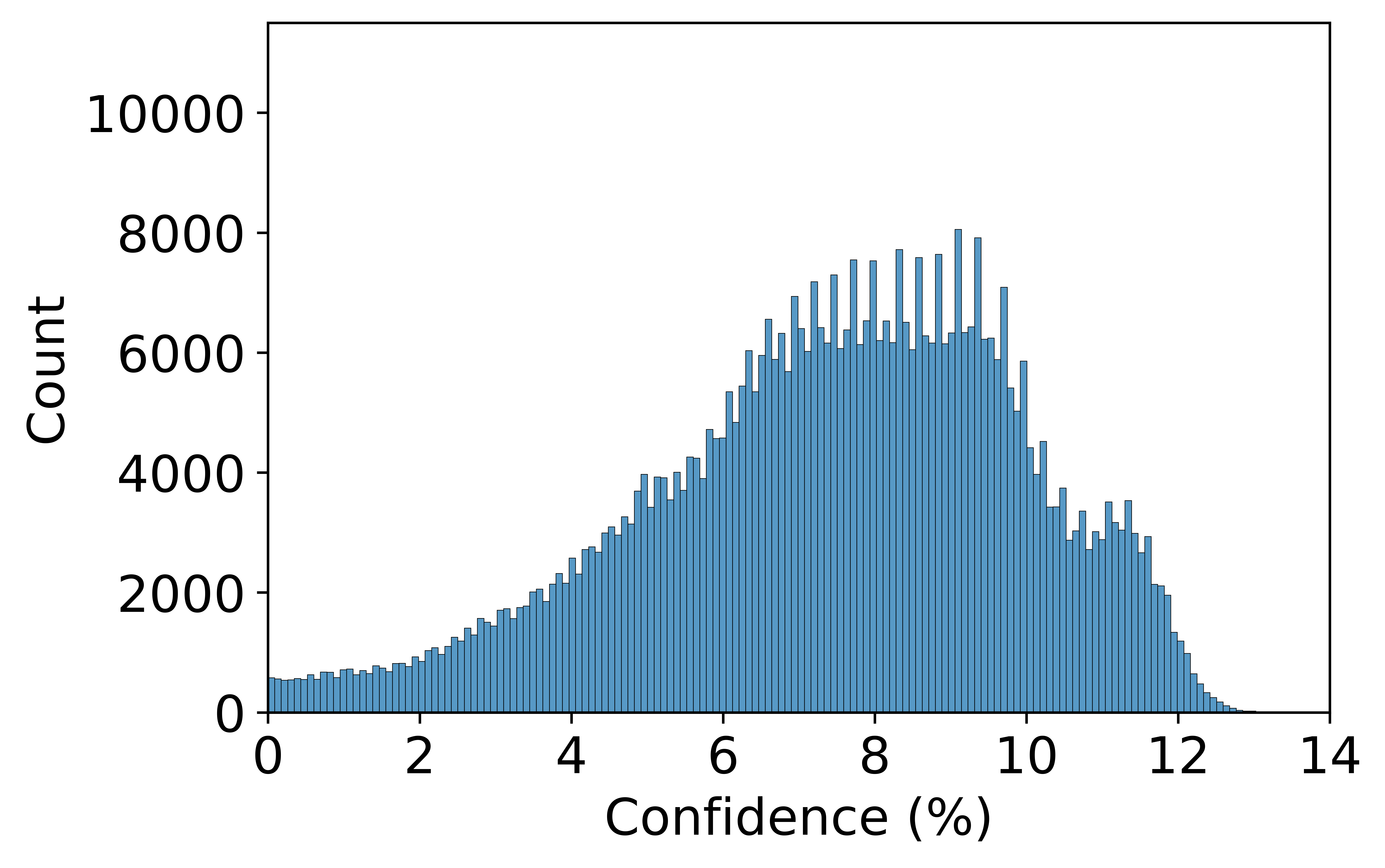}
    \caption{HAND \\ $\alpha = 5.00$.}
    \label{fig:confHAND5.00}
\end{subfigure}
\caption{Distribution of the confidence values of all correctly classified training samples of UCIHAR (first column), CTG (second column), ISOLET (third column) and HAND (last column) after training with $\alpha = 0.00$ (first row) and with $\alpha$ yielding the best accuracy for the considered dataset (second row).}
\label{fig:confafter}
\end{figure}

\noindent The effect of introducing the confidence threshold in the training procedure of HDC is visualized in Figure \ref{fig:confafter}. This figure gives for each of the four datasets (one column for each) the distribution of the confidence values of all correctly classified training samples after training with $\alpha = 0.00$ (baseline, top row) and after training with the setting of $\alpha$ resulting in the best performance for the considered dataset (bottom row). When comparing with Figure \ref{fig:conf}, already larger confidence values are seen after training with $\alpha = 0.00$ for all four datasets. However, there is an even more clear shift in the distribution towards higher confidence values for the best settings of $\alpha$ for each dataset illustrating nicely the effect of the proposed simple extended training procedure in the HDC framework. \\

\noindent The abovementioned results show the impact of the parameter $\alpha$ on the test set. To determine the optimal $\alpha$ for a particular dataset, the standard procedure of using a validation set should be used. Table \ref{tab:resultsCV} in the appendix illustrates the procedure for the ISOLET dataset when performing 10-fold cross validation, resulting in an optimal $\alpha$ of 1.25. \\

\noindent The obtained highest accuracy of 94.30\% for ISOLET is an improvement compared to the accuracy of 92.4\% obtained by \textcite{Imani2019a} using an adaptive learning rate during training procedure and by \textcite{Hsiao2021} who project original HDC to learnable HDC that is trained similarly as a BNN. Also for the CTG dataset, the highest accuracy of 86.89\% is higher compared to the accuracy of 82\% obtained by \textcite{Basaklar2021} who flip a variable number of bits between levels when applying \textit{linear mapping}. \\

\noindent While our method consistently improves the accuracy compared to baseline, some other studies in literature report slightly better results on the considered datasets but with often more complex methods. For example, \textcite{Imani2019a} report an accuracy of 96\% on the UCIHAR dataset with their adaptive learning rate, \textcite{Hernandez2021a} use weighted bundling in and out of class bundles and obtain an accuracy of 96.5\% for UCIHAR and 94.6\% for ISOLET, learnable HDC of \textcite{Hsiao2021} results in an accuracy of 95.54\% for UCIHAR, \textcite{Zou2021a} obtain an accuracy of 98\% for UCIHAR and 95\% for ISOLET applying manifold learning and \textcite{Duan2022a} map HDC into a BNN achieving 95.23\% and 94.89\% accuracy for UCIHAR and ISOLET, respectively. \\

\noindent As future work, the proposed training procedure could possibly be extended with an adaptive confidence threshold. Namely, the distribution of confidence values shift towards higher confidence while correcting for low-confident correctly classified training samples (Figure \ref{fig:confafter}) such that the threshold could increase along with this shift in distribution. Moreover, it might be interesting to combine the proposed extended training procedure with previously proposed adjustments to the training procedure (\cite{Imani2019a, Hernandez2021a}) and investigate whether this would further improve the HDC performance.

\section*{Conclusion}

A simple, yet effective extension to the training procedure in binary HDC is introduced which takes into account not only wrongly classified samples, but also samples that are correctly classified by HDC but with a low confidence. A threshold on the confidence value is introduced and tested on four datasets for which the performance consistently improves compared to the baseline across a range of confidence threshold values. The extended training procedure also results in a shift towards higher confidence values of the correctly classified samples making the classifier not only more accurate but also more confident about its predictions.

\subsection*{Acknowledgments}
This research received funding from the Flemish Government under the "Onderzoeksprogramma Artifici\"ele Intelligentie (AI) Vlaanderen" programme.

\section*{Appendix}
\setcounter{section}{0}
\renewcommand{\thesection}{A.\arabic{section}}
\setcounter{table}{0}
\renewcommand{\thetable}{A\arabic{table}}

\section{Notation}

A summary of notation can be found in Table \ref{tab:symbols}.

\begin{table}[!hbt]
\centering
\caption{List of symbols. (HD = hyperdimensional, CIM = Continuous Item Memory)}
\label{tab:symbols}
\begin{tabular}{@{}rl|rl@{}}
\hline
 \textbf{Symbol} & \textbf{Definition} & \textbf{Symbol} & \textbf{Definition} \\
\hline
$x$      & vector in input space &               &                                            \\
$m$      & number of samples     & $i$           & $1...m$                                    \\
$n$      & number of features    & $j$           & $1...n$                                    \\
$K$      & number of classes     & $k$           & $1...K$                                    \\
$D$      & HD vector dimension   & $d$           & $1...D$                                    \\
$s$      & similarity            & $y_{i}$       & true class of $i$th sample                 \\
$h$      & Hamming distance      & $\hat{y}_{i}$ & predicted class of $i$th sample            \\
$c$      & confidence value      & $l$           & true class label                           \\
$\alpha$ & confidence threshold  & $\hat{l}$     & predicted class label                      \\
         &                       & $\hat{l}'$    & class label with second highest similarity \\
 & & & \\
$\textbf{v}$ & vector in HD space $\mathcal{H}$ & $\textbf{B}$ & bundle in HD space $\mathcal{B}$ \\
$\textbf{s}$ & sample vector                    & $\textbf{S}$ & sample bundle                    \\
$\textbf{c}$ & class vector/prototype           & $\textbf{C}$ & class bundle                     \\
$\textbf{q}$ & query vector                     &                                                 \\
 & & & \\
$\mathcal{H}$ & vector HD space, $\{0,1\}^{D}$ & $\mathcal{B}$ & bundle HD space, $\mathbb{N}^{D}$ \\
$CIM_{j}$     & CIM of $j$th feature           & $\oplus$      & bundling operator                 \\
$\otimes$     & binding operator               & $\ominus$     & bundling out operator             \\
$\rho$        & permutation operator           & $[.]$         & majority rule                     \\ \hline
\end{tabular}
\end{table}

\section{Cross-Validation}

Table \ref{tab:resultsCV} contains the results of 10-fold cross validation (CV) on the training set of ISOLET for the different settings of $\alpha$. The table includes the training accuracy, validation accuracy and validation error rate, averaged over the ten folds of 10-fold CV. This shows that the optimal $\alpha$ value with 10-fold CV is $\alpha = 1.25$, resulting in a test accuracy of 93.98\% (Table \ref{tab:resultsISOLET}) which is very near to the best test accuracy achieved in Table \ref{tab:resultsISOLET}, i.e., 94.30\% for $\alpha = 1.00$.

\begin{table}[!hbt]
\centering
\caption{Averaged accuracy (\%) on the train and validation folds and averaged error rate (\%) on the validation folds of 10-fold cross validation for ISOLET for the tested settings of the confidence threshold $\alpha$. Data are \textit{mean} ($\pm$ \textit{standard deviation}), in \textbf{bold} are the validation accuracies (and error rates) that are higher (and lower) than the baseline ($\alpha = 0.00$) and \underline{underlined} is the best validation accuracy and error rate.}
\label{tab:resultsCV}
\begin{tabular}{@{}cccc@{}}
    \hline
    \textbf{$\alpha$} & \textbf{Train accuracy} & \textbf{Validation accuracy} & \textbf{Validation error rate} \\
    \hline
    0.00 & 100.00 ($\pm$ 0.00) & 90.56 ($\pm$ 1.97) & 9.44 \\
    0.25 & 100.00 ($\pm$ 0.00) & \textbf{92.43} ($\pm$ 1.96) & \textbf{7.57} \\
    0.50 & 99.99 ($\pm$ 0.01) & \textbf{93.14} ($\pm$ 1.85) & \textbf{6.86} \\
    0.75 & 99.92 ($\pm$ 0.13) & \textbf{93.64} ($\pm$ 1.30) & \textbf{6.36} \\
    1.00 & 99.62 ($\pm$ 0.29) & \textbf{93.41} ($\pm$ 2.06) & \textbf{6.59} \\
    1.25 & 99.78 ($\pm$ 0.21) & \underline{\textbf{93.83}} ($\pm$ 1.93) & \underline{\textbf{6.17}} \\
    1.50 & 99.61 ($\pm$ 0.24) & \textbf{93.68} ($\pm$ 1.55) & \textbf{6.32} \\ \hline
\end{tabular}
\end{table}

\clearpage
\printbibliography





\end{document}